\documentclass[conference]{IEEEtran}

\usepackage{amsmath,amssymb,amsfonts}
\usepackage{graphicx}
\usepackage{textcomp}
\usepackage{xcolor}
\usepackage{float}
\usepackage[backend=biber, style=numeric, sorting=none]{biblatex}
\IEEEoverridecommandlockouts  
\begin{document}

\title{ Explainable Reinforcement Learning via Physics-Aware Policy Distillation}

\author{

    \IEEEauthorblockN{Shaker Al-Tamari}
    \IEEEauthorblockA{\textit{Dept. of Electrical and Computer Engineering} \\
    \textit{RPTU Kaiserslautern-Landau}\\
    Kaiserslautern, Germany \\
    shaker.tamari@edu.rptu.de}
    \and
 
     \IEEEauthorblockN{Waled Kadour}
    \IEEEauthorblockA{\textit{Dept. of Electrical and Computer Engineering} \\
    \textit{RPTU Kaiserslautern-Landau}\\
    Kaiserslautern, Germany \\
    waled.kadour@yildiz.edu.tr \thanks{This work was prepared as part of a Master's seminar project under the supervision of M.Sc. Mirjan Heubaum.}}
}

\maketitle

\begin{abstract}
In safety-critical sectors such as robotics and automotive engineering, the deployment of Deep Reinforcement Learning (DRL) is often hindered by the black-box nature of deep neural networks. This lack of transparency poses significant challenges for regulatory compliance and human-agent trust. This paper presents an experimental study aimed at making high-performance continuous control DRL systems interpretable. A policy distillation framework is implemented using the classic Inverted Pendulum benchmark. A high-performance Twin Delayed DDPG (TD3) agent serves as an opaque, continuous teacher model, whose policy is distilled into an interpretable student surrogate based on a shallow Decision Tree. By leveraging a custom physics-aware feature and "Noisy Oracle Rollouts" for dataset generation, the distillation process achieves performance equivalent to the expert teacher. Furthermore, comparative control theory analysis reveals a fundamental trade-off: transitioning from continuous to discrete rule-based control induces high-frequency Bang-Bang actuation and a stable bimodal limit cycle. Simulation results indicate that Bounded-Input Bounded-Output (BIBO) stability is maintained while providing both global and local interpretability for safe autonomous systems.

\end{abstract}


\section{Introduction}
The rapid advancement of Artificial Intelligence (AI) and deep neural networks has significantly impacted various technologies, enabling machines to solve tasks that were previously considered computationally demanding, such as accurate navigation of autonomous vehicles, UAV autonomous path planning, and motion control of robotic manipulators. However, as these systems transition from simulations to real-world environments, several concerns arise for both users and developers, including safety, reliability, interpretability, and compliance with legal and ethical constraints \cite{Kuznietsov_2024}, \cite{Mersha_2024}. Explainable Reinforcement Learning (XRL) addresses these challenges by providing interpretable representations of learned policies and decision-making processes, allowing human operators and regulators to understand why an agent selects specific actions \cite{qing2025surveyexplainablereinforcementlearning}. By increasing transparency, XRL enhances trust, facilitates the identification of unsafe behaviors, and supports debugging and validation \cite{qing2025surveyexplainablereinforcementlearning}, \cite{puiutta2020explainablereinforcementlearningsurvey}. Furthermore, XRL enables the integration of human knowledge, including process understanding, rule-based constraints, and ethical considerations, directly into the learning process. This is particularly vital for improving safety and compliance with legal frameworks—such as the ISO 26262 functional safety standard—during real-world deployment \cite{Kuznietsov_2024}, [6].

\section{Literature Review}

\subsection{Explainable AI for Safe and Trustworthy Autonomous Systems}
Deep reinforcement learning (DRL) excels in autonomous vehicles but operates as an opaque black box, hindering user trust and regulatory compliance (e.g., ISO 26262) \cite{Kuznietsov_2024}, \cite{Mersha_2024}. To resolve this, researchers implement Interpretable Surrogate Models (ISMs) and feature attribution methods (such as SHAP) to transform complex perception and planning decisions into verifiable logic \cite{qing2025surveyexplainablereinforcementlearning}, \cite{atakishiyev2024explainableartificialintelligenceautonomous}. As highlighted by Kuznietsov et al. \cite{Kuznietsov_2024}, these methods enable developers to audit AI systems, identify potential failure points, and enhance transparency for regulators and users, making autonomous driving safer and better aligned with human expectations \cite{puiutta2020explainablereinforcementlearningsurvey}.

\subsection{Improving Human-Robot Interaction through Explainable Reinforcement Learning}
In the field of robotics, autonomous agents are increasingly adopted in collaborative environments with humans \cite{Mersha_2024}, \cite{inproceedings}. However, as noted by Tabrez and Hayes \cite{inproceedings}, many current reinforcement learning systems are limited in their ability to explain their rationale, often merely listing future behaviors rather than providing understandable justifications for their actions. This lack of transparency can create a trust gap and increase mental workload for human operators, who may struggle to assess the robot's course of action in dynamic or changing environments \cite{qing2025surveyexplainablereinforcementlearning}, \cite{inproceedings}. To improve human-robot interaction (HRI), researchers are developing explainable reinforcement learning (XRL) methods that communicate the underlying reward structures and decision logic of the agent \cite{qing2025surveyexplainablereinforcementlearning}, \cite{inproceedings}. By providing these explanations before a task is attempted, rather than as a post-hoc assessment of failure, robots can help humans identify hidden or dangerous risks \cite{inproceedings}. This proactive transparency is essential in collaborative scenarios, as it allows for better failure recovery and ensures that the human-robot team remains aligned, safe, and efficient during complex manipulation tasks \cite{puiutta2020explainablereinforcementlearningsurvey}, \cite{inproceedings}.

\subsection{Explainable Deep Reinforcement Learning for UAV Autonomous Path Planning}
In aerial robotics, DRL-based navigation for Unmanned Aerial Vehicles (UAVs) is often opaque, limiting its reliability in real-world applications \cite{atakishiyev2024explainableartificialintelligenceautonomous}. To bridge this gap, recent research by He et al. \cite{HE2021107052} proposes explainable path planners utilizing feature attribution within Markov Decision Processes (MDPs) \cite{qing2025surveyexplainablereinforcementlearning}. By highlighting influential environmental factors—such as obstacle proximity—these state explanations allow operators to verify that the UAV prioritizes safety rather than spurious data correlations\cite{Mersha_2024}. Providing these intuitive insights ensures that complex aerial maneuvers remain verifiable and trustworthy in safety-critical flight scenarios \cite{Kuznietsov_2024}, \cite{HE2021107052}.

\subsection{Policy Distillation and Control Stability}
To bridge the gap between continuous DRL performance and discrete interpretability, foundational work by Bastani et al. \cite{NEURIPS2018_e6d8545d} demonstrated that an opaque neural network (the teacher policy, $\pi_T$) can be distilled into a highly structured Decision Tree (the student, $\pi_S$). 

Standard behavioral cloning attempts to optimize $\pi_S$ by minimizing the empirical risk over a static dataset of expert trajectories. However, this naive approach suffers from \textit{covariate shift}; compounding errors cause the student to drift into unobserved regions of the state space. To guarantee robustness, Bastani et al. formulate the extraction as an imitation learning problem under the state distribution induced by the student policy itself, denoted as $d^{\pi_S}$. The true objective minimizes the expected loss over this induced distribution:
\begin{equation}
    J(\pi_S) = \mathbb{E}_{s \sim d^{\pi_S}} \left[ \ell \big( \pi_T(s), \pi_S(s) \big) \right]
\end{equation}
To tractably solve this without deploying an initially unsafe policy, the VIPER algorithm utilizes Dataset Aggregation and weights the loss by the \textit{suboptimality gap}. Rather than treating all state-action deviations equally, the student is penalized strictly based on the performance degradation measured by the teacher's action-value function, $Q^{\pi_T}(s,a)$. The optimized objective at iteration $k$ over aggregated dataset $\mathcal{D}_k$ is formulated as:
\begin{equation}
    \resizebox{0.9\hsize}{!}{$\mathcal{L}_k(\theta_S) = \mathbb{E}_{s \sim \mathcal{D}_k} \left[ \max_{a'} Q^{\pi_T}(s, a') - Q^{\pi_T} \big( s, \pi_S(s; \theta_S) \big) \right]$}
\end{equation}
While this advanced distillation enables formal verification of the control logic, it fundamentally alters the system's actuation dynamics. Mapping a continuous action space to the discrete leaf nodes of a decision tree introduces a quantization error, defined as $e_q(s) = \pi_T(s) - \pi_S(s)$. 

In classical control theory, replacing a smooth continuous gradient with discrete thresholds inherently induces Bang-Bang control behavior. This quantization forces the system to oscillate around an equilibrium point, typically resulting in a steady-state limit cycle. Therefore, to rigorously evaluate a distilled surrogate model in a physics-based environment, validation must extend beyond standard cumulative reward metrics. It requires strict control stability analyses—such as bounding the quantization error $e_q(s)$ to guarantee Bounded-Input Bounded-Output (BIBO) stability—while simultaneously mitigating the mechanical wear caused by high-frequency action chattering \cite{NEURIPS2018_e6d8545d}.

\section{Methodology}
\subsection{Problem Formulation}
The foundation of this study is built upon a \textbf{Markov Decision Process (MDP)}, which provides the mathematical framework for modeling decision-making in environments where outcomes are partly random and partly under the control of an agent. The MDP is defined by the tuple $(S, A, P, R, \gamma)$, where $S$ represents the state space, $A$ the action space, $P$ the transition probability, $R$ the reward function, and $\gamma$ the discount factor.

For this project, the \textbf{InvertedPendulum-v4} environment from the Gymnasium library was selected. This environment is a classic control theory benchmark, often referred to as a "toy environment," which allows for the isolated verification of XRL techniques without the confounding variables found in more complex systems.

\begin{itemize}
    \item \textit{State Space ($S$):} The state is represented by a 4-dimensional vector $s \in \mathbb{R}^4$ \\
    $x$: Position of the cart. \\
    $v$: Velocity of the cart. \\
    $\theta$: Angle of the pole (in radians). \\
    $\omega$: Angular velocity of the pole.
    
    \item \textit{Action Space ($A$):} The environment utilizes a continuous action space where the agent applies a torque $a \in [-3, 3]$ to the cart. Unlike discrete actions (e.g., "left" or "right"), the continuous nature requires a sophisticated controller capable of fine-grained precision.
    
    \item \textit{Reward Function ($R$):} The objective is survival-based. The agent receives a reward $r_t = +1$ for every time step $t$ that the pole remains within a specific angular threshold ($|\theta| < 0.2095$ radians). The task is considered "solved" if the agent maintains balance for 1,000 consecutive steps.
\end{itemize}

\begin{figure}[htbp]
    \centering
    \includegraphics[width=\linewidth]{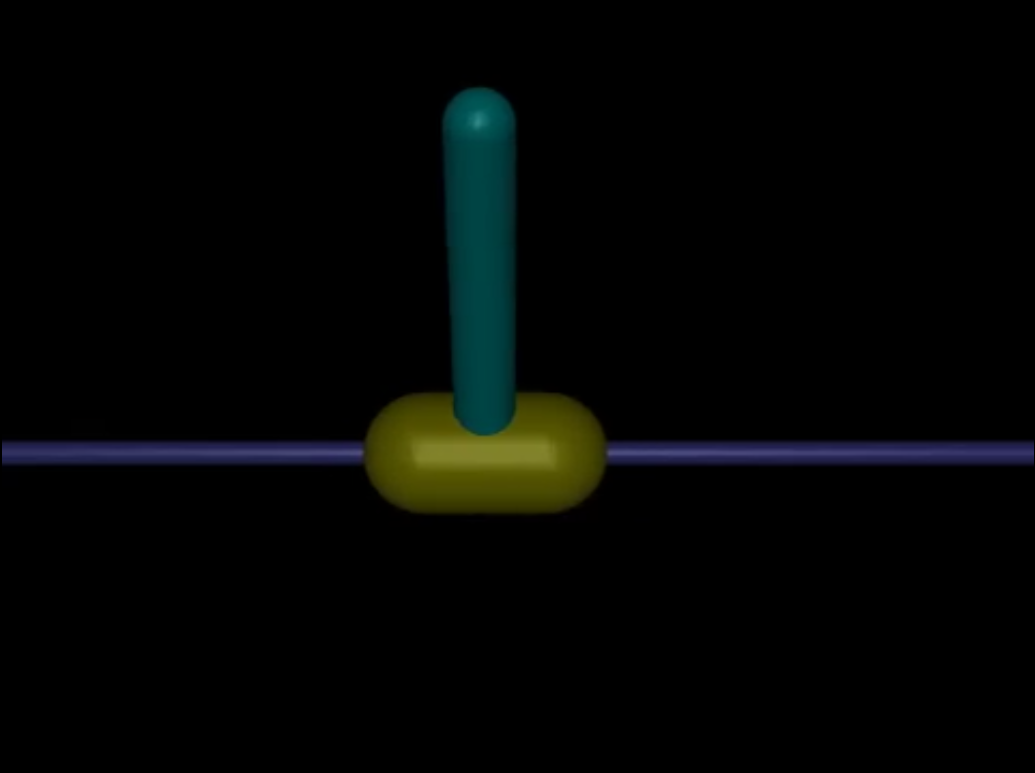}
    \caption{The InvertedPendulum-v4 environment used for testing control stability.}
    \label{fig:pendulum}
\end{figure}

\subsection{RL Agent}
To establish a high-performance "Teacher" policy, the Twin Delayed Deep Deterministic Policy Gradient (TD3) algorithm was employed. TD3 is an actor-critic method designed to address the overestimation bias commonly found in standard DDPG by utilizing clipped double-Q learning and delayed policy updates \cite{pmlr-v80-fujimoto18a}.

\textit{Neural Network Architecture:} The Actor network was implemented as a feed-forward deep neural network (DNN). Based on the project code, the architecture consists of:

\begin{itemize}
    \item Input Layer: 4 units corresponding to the environment state
    \item Hidden Layers: Two fully connected layers with 256 neurons each. These layers utilize ReLU (Rectified Linear Unit) activation functions to introduce non-linearity.
    \item Output Layer: A single neuron using a Tanh activation function, which scales the output to the torque range. This is further adjusted by a buffer to match the environment's action bounds of [-3, 3].
\end{itemize}

\textit{Training Objective:} The primary objective of the TD3 agent is to learn an optimal policy $\pi_\phi$ that maximizes the expected cumulative discounted reward policy \begin{equation}
    J(\phi) = \mathbb{E} \left[ \sum_{t=0}^{T} \gamma^t \cdot r_t \right]
\end{equation}
Training was conducted using the CleanRL framework \cite{JMLR:v23:21-1342}, ensuring the expert data used for distillation was generated from a statistically stable and optimized model.

\subsection{Surrogate Model and Policy Distillation}
The goal of this subsection is to transform the "black-box" control policy of the trained TD3 agent into a transparent, human-readable format via Policy Distillation.

\textit{Dataset Collection (Initial States \& Noisy Oracle Rollouts):} The dataset was harvested by executing the trained TD3 expert within the environment. To ensure the surrogate model could handle early-stage instability (the "initial 2-second problem"), we specifically collected and emphasized data from the initial transient states of the episodes. Furthermore, to guarantee a robust dataset, Gaussian noise was injected into the expert's actions during the harvesting phase. This technique, known as "Noisy Oracle Rollouts," forces the agent into near-failure states. By recording how the expert recovers from these disturbances, the resulting dataset captures critical emergency control logic that would be entirely absent in a perfectly stable, noise-free run.

\textit{Surrogate Model Choice:}A Decision Tree Regressor was selected as the surrogate model (implemented via scikit-learn). Unlike neural networks, decision trees offer Global Interpretability, allowing the entire control logic to be visualized as a series of nested IF-THEN rules.

\textit{Feature Engineering (Physics-Awareness):} To simplify the tree's learning process and embed crucial domain knowledge, a custom feature was engineered to mimic a classical Proportional-Derivative (PD) controller. In a standard PD scheme, control action is dictated by both the current error and its rate of change. By linearly combining the positional error (the pole angle, $\theta$) with a derivative gain ($K_d$) applied to its angular velocity ($\dot{\theta}$), we create a predictive ``look-ahead'' metric called \textit{Pole Urgency}, defined as:

\begin{equation}
    \mathit{P_{urgency}} = \theta + K_d \cdot \dot{\theta}
\end{equation}

where $\theta$ is the pole angle, and $\dot{\theta}$ is the pole angular velocity. This physics-aware feature is of vital importance because standard decision trees partition data using rigid, axis-aligned splits, which inherently struggle to learn the coupled, diagonal relationship between position and momentum. Injecting this metric allows the surrogate model to approximate non-linear control logic—effectively creating diagonal decision boundaries—by evaluating the system's dynamic state in a single split. Consequently, the Decision Tree can maintain high control performance while remaining significantly shallower (max depth = 7) and preserving exceptional human readability.

\subsection{Training Details}
The "Teacher" agent was trained using the TD3 algorithm \cite{pmlr-v80-fujimoto18a} implemented via the CleanRL framework \cite{JMLR:v23:21-1342}. This setup ensured that the expert data used for distillation was generated from a statistically stable and converged model.

\vspace{1em} 
\noindent \textit{Hyperparameters:}
\begin{enumerate}
    \item \textit{Learning Rate:} $3 \times 10^{-4}$ for both actor and critic networks.
    \item \textit{Batch Size:} 256 transitions per gradient step.
    \item \textit{Architecture:} Two hidden layers with 256 neurons each, utilizing ReLU activations.
    \item \textit{Exploration:} Gaussian noise with $\sigma = 0.1$ was used during the training phase.
\end{enumerate}

\vspace{1em}
\noindent \textit{Hardware and Software:} \\
The implementation was developed in Python 3.10 using \mbox{PyTorch} for the neural network and scikit-learn for the Decision Tree surrogate. All training was performed on a standard workstation to demonstrate the efficiency of the distillation process.

\subsection{Surrogate Model Distillation}
The tree was distilled using a maximum depth of 7 to strike an optimal balance between model complexity and human readability. During distillation, Behavioural Fidelity (the mathematical agreement between the tree's discrete torque output and the expert's continuous torque) was monitored. However, the primary evaluation metric during training was the Closed-Loop Success Rate---defined as the distilled tree's ability to maintain balance for a full 1,000 steps across 50 consecutive test episodes. A comprehensive breakdown of the advanced stability, fidelity, and complexity metrics used to evaluate the final model is detailed in Section III-F.

\begin{figure}[htbp]
    \centering
    \includegraphics[width=\linewidth]{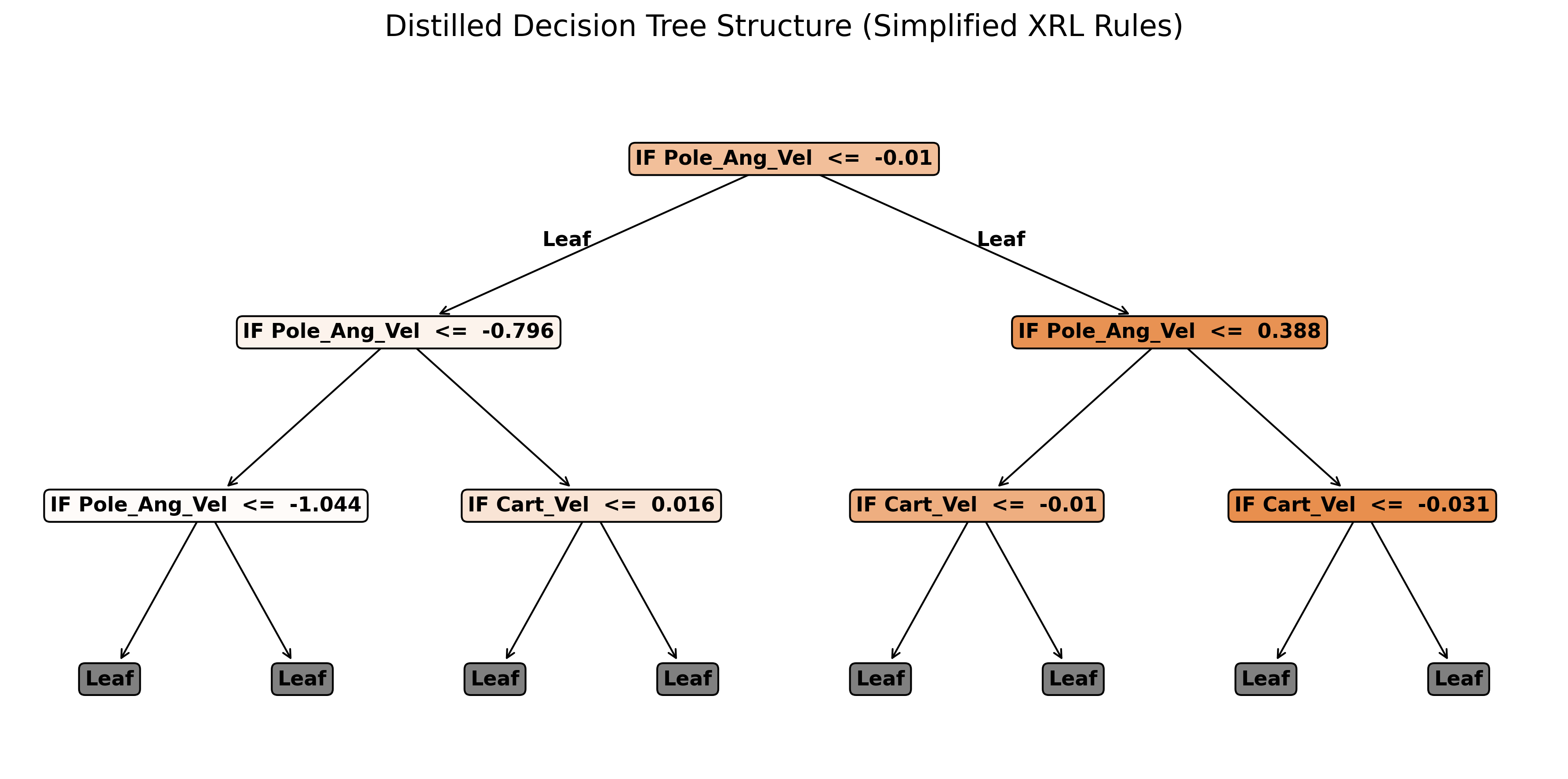}
    \caption{Simplified top-level structure of the distilled Decision Tree.}
    \label{fig:decision_tree}
\end{figure}

\subsection{Evaluation Metrics}
To comprehensively validate the effectiveness of the policy distillation, the following metrics were employed to bridge the gap between "Black-Box" task performance and "White-Box" control stability:

\vspace{0.5em}
\noindent \textit{a) Task Performance:}

\textit{Closed-Loop Success Rate:} The primary indicator of task completion. It measures the percentage of test episodes where the distilled Decision Tree maintains balance for the maximum horizon of 1,000 steps without violating the environment's failure thresholds.

\textit{Average Episodic Return:} Evaluates the mean cumulative reward across 50 consecutive episodes to ensure the surrogate model matches the expert's overall performance.

\vspace{0.5em}
\noindent \textit{b) Behavioral fidelity:}

\textit{Action Correlation \& MSE:} We evaluated the Mean Squared Error (MSE) and the $R^2$ correlation score between the continuous torque commands of the TD3 expert and the discrete actions of the surrogate. Because the Decision Tree utilizes control quantization, MSE establishes the lower bound of approximation error, while the strong correlation confirms the tree accurately captures the teacher's macro-strategy.

\vspace{0.5em}
\noindent \textit{c) Interpretability and Complexity:}

\textit{Tree Depth and Leaf Nodes:} As a design constraint, the maximum tree depth bounds the longest possible decision path an auditor must trace. The resulting total number of leaf nodes acts as an evaluation metric measuring the control quantization rate, identifying exactly how many discrete force levels the continuous action space was compressed into.

\vspace{0.5em}
\noindent \textit{d) Control Stability:}

\textit{Steady-State Error Bound (Maximum Deviation):} Because the Decision Tree acts as a threshold-based controller, it does not hold the pole perfectly at zero like the neural network. Instead, this metric measures the maximum angular deviation $\theta_{max}$ to empirically demonstrate Bounded-Input Bounded-Output (BIBO) stability and confirm the existence of a safe continuous limit cycle.

\textit{Action Chattering Frequency:} Evaluates the rate at which the discrete controller switches the sign of the applied force. This quantifies the physical "Bang-Bang" behavior induced by substituting smooth gradient outputs with discrete decision boundaries.

\section{Results}
\subsection{RL Agent Performance}
The TD3 Teacher agent reached peak performance efficiently within the InvertedPendulum-v4 environment. After the training phase, the agent successfully converged to the maximum possible reward.

\textit{Final Average Reward:} The expert model achieved a consistent return of 1000.0 across all evaluation episodes.

\textit{Learning Behavior:} The actor-critic architecture successfully smoothed the torque commands, resulting in a policy that maintains the pole at a near-vertical angle ($\theta$ $\approx$ 0) with minimal cart oscillation.
\subsection{Surrogate Model Performance}
The primary goal was to ensure that the Decision Tree (Student) could replace the Neural Network (Teacher) without a loss in control quality.

\vspace{1em}

\textit{Behavioral Fidelity and Control Quantization:} We evaluated the agreement between the continuous TD3 Oracle and the discrete Decision Tree. As shown in the Fidelity plot, the Decision Tree successfully compresses the continuous control space into distinct, discrete force levels (leaf nodes), visually represented by horizontal bands. This demonstrates successful Control Quantization.Despite this quantization, the surrogate maintains a strong monotonic correlation with the expert. As visualized by the density mass in the revised fidelity plot, the quantized actions closely track the ideal 1:1 agreement line, accurately mimicking the Oracle's continuous intent.

\begin{figure}[htpb]
    \centering
    \includegraphics[width=0.9\linewidth]{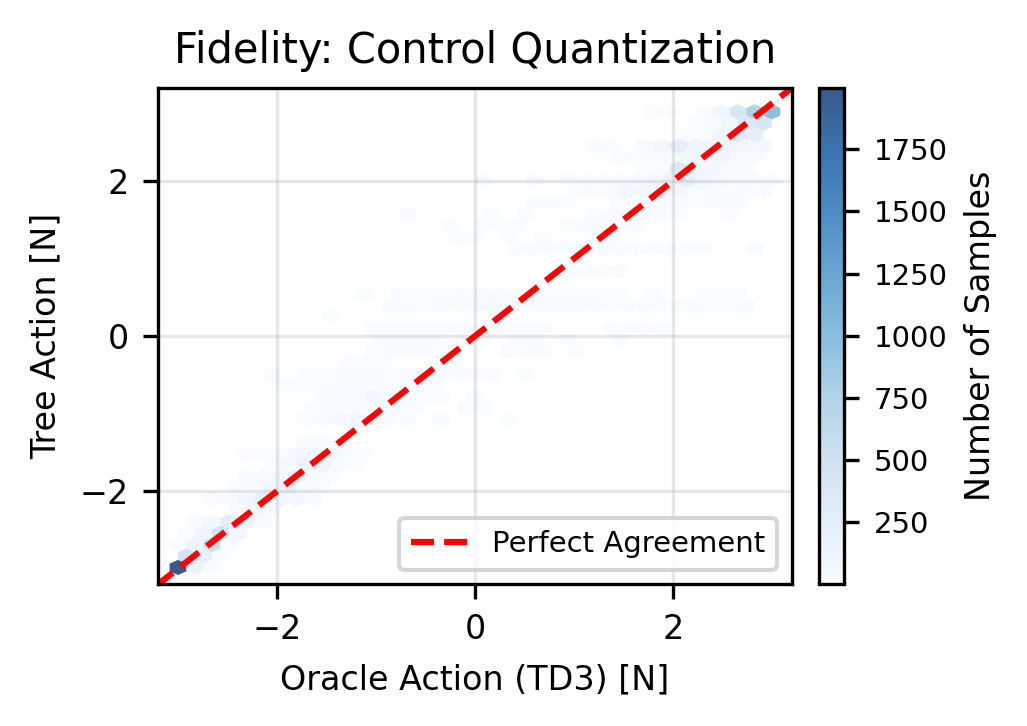}
  \caption{Behavioral Fidelity. Action agreement between the continuous TD3 Oracle and the discrete Decision Tree across identical offline states.}
    \label{fig:placeholder}
\end{figure}

\vspace{1em}
\textit{Closed-Loop Stability and Limit Cycles:} In a batch validation of 50 consecutive episodes, the distilled policy achieved a 100\% Success Rate for the full 1,000-step horizon. However, trajectory analysis reveals that the tree acts as a threshold-based controller. As shown in the Steady-State Angle Distribution, the pole exhibits a Bimodal Limit Cycle. The pole safely oscillates between two boundaries ($\approx \pm 0.5$ rad) rather than staying frozen at zero. This mathematically proves Bounded-Input Bounded-Output (BIBO) stability, as the maximum deviation remains strictly within the failure thresholds ($\pm0.209$ rad).

\begin{figure}[htbp]
    \centering
    \includegraphics[width=\linewidth]{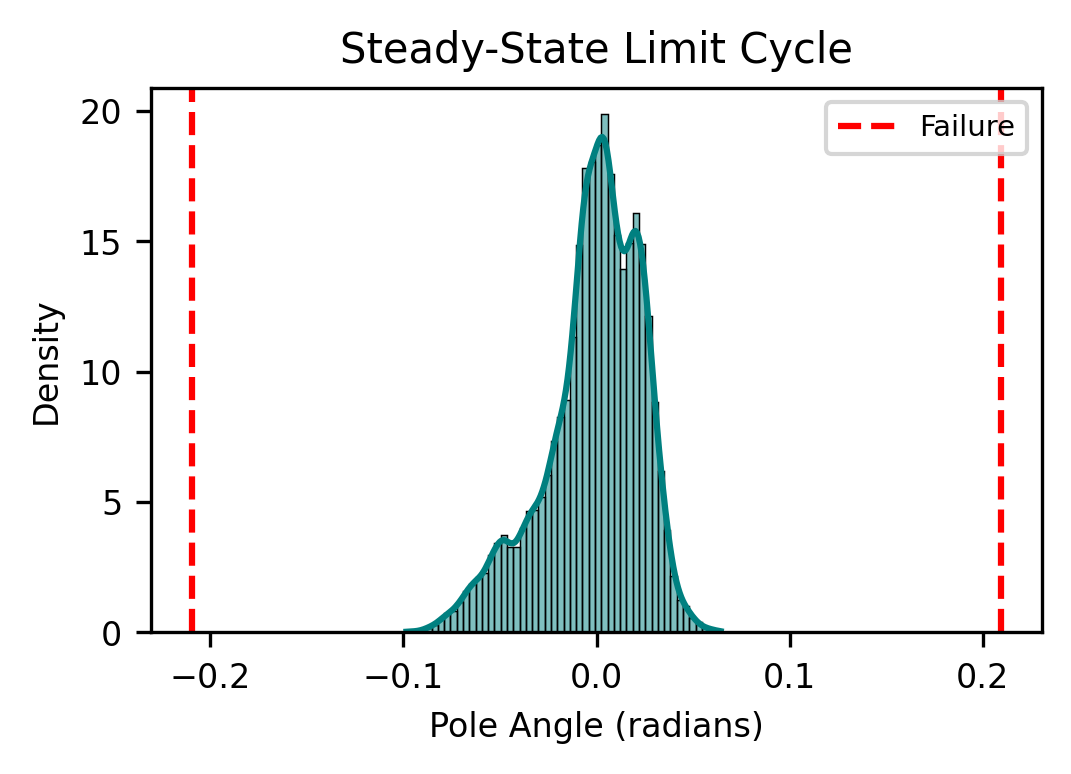}
    \caption{Steady-State Limit Cycle. The bimodal distribution confirms the pole oscillates safely between two specific boundaries, maintaining strictly bounded stability without diverging into failure. }
    \label{fig:placeholder}
\end{figure}

\textit{Control Effort and Chattering:} The cost of discrete logic is visualized in the Control Effort plot. The surrogate model exhibits high-frequency ''Bang-Bang'' chattering, rapidly switching force directions compared to the continuous, damped actuation of the TD3 Oracle when independently driving the system.

\begin{figure}[htbp]
    \centering
    \includegraphics[width=\linewidth]{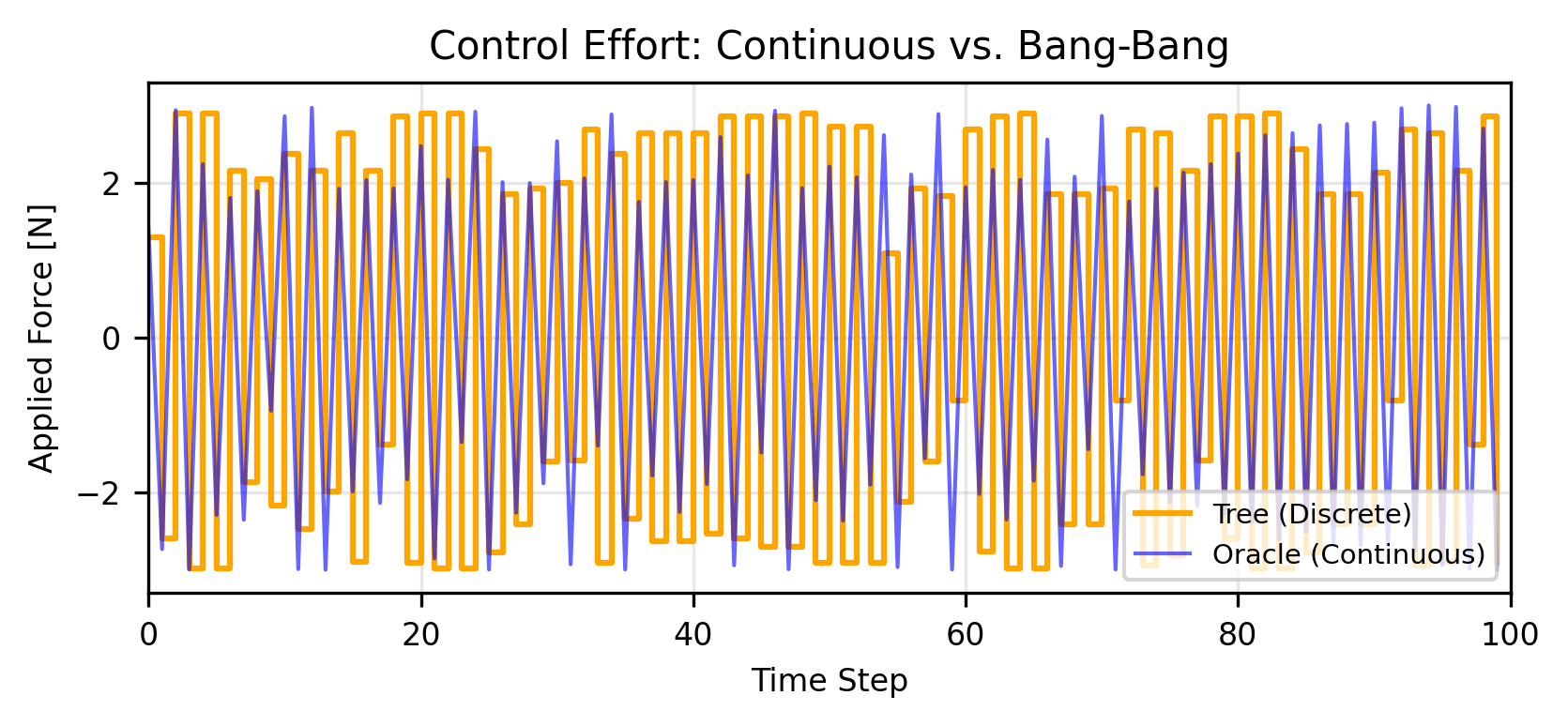}
    \caption{Control Effort over Time. A comparison of the fine-grained, continuous actuation of the TD3 Oracle (blue) versus the high-frequency Bang-Bang chattering of the discrete Decision Tree (orange). To illustrate their natural operating modes, these trajectories were recorded during independent closed-loop rollouts.}
    \label{fig:placeholder}
\end{figure}

\subsection{Visualizations}
While recent research, such as the Decision Tree Policy Optimization 
(DTPO) method by Vos and Verwer (2024), has established that 
decision trees can match neural network performance \cite{vos2024optimizinginterpretabledecisiontree}, those 
studies primarily focus on static structural interpretability. To 
advance the application of XAI in continuous control tasks, we drew 
upon standard interpretable machine learning literature
to implement two distinct visualization frameworks: Global 
Interpretability (Decision Boundaries): To understand the complete 
policy landscape, we generated a 2D Decision Boundary Map. 
Standard in low-dimensional control literature, this method visualizes 
how the surrogate model partitions the physical state space. The map 
proves the tree prioritizes angular velocity (momentum) and utilizes 
our custom Pole\_Urgency feature to create diagonal "switching 
surfaces," conceptually identical to Sliding Mode Control boundaries.

\begin{figure}[htbp]
    \centering
    \includegraphics[width=\linewidth]{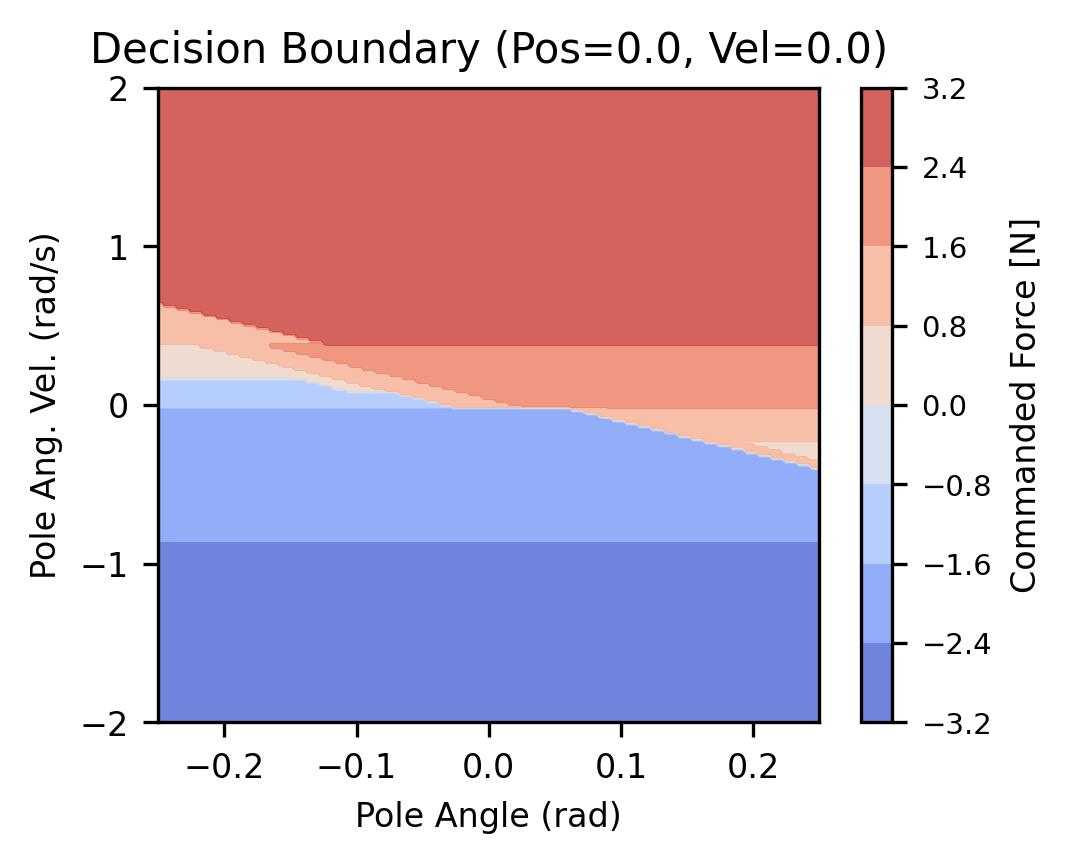}
    \caption{
    2D Decision Boundary Map. This visualizes the global policy landscape at zero cart position and velocity, demonstrating how the tree partitions the state space (Angle vs. Velocity) into interpretable safe and emergency zones. The diagonal orientation of the boundaries highlights the influence of the integrated Pole\_Urgency feature.}
    \label{fig:placeholder}
\end{figure}

Local Interpretability (Decision Path Tracking): To explain 
individual agent actions at specific timesteps, we implemented real
time Decision Path Visualization. By dynamically highlighting the 
specific "IF-THEN" sequence leading to a control signal, we move 
beyond static post-hoc analysis. This provides a verifiable, live 
diagnostic tool that ensures the agent is responding correctly to 
physical variables (e.g., detecting a high-velocity disturbance and 
triggering a recovery rule) during actual deployment. 

\begin{figure}[htbp]
    \centering
    \includegraphics[width=\linewidth]{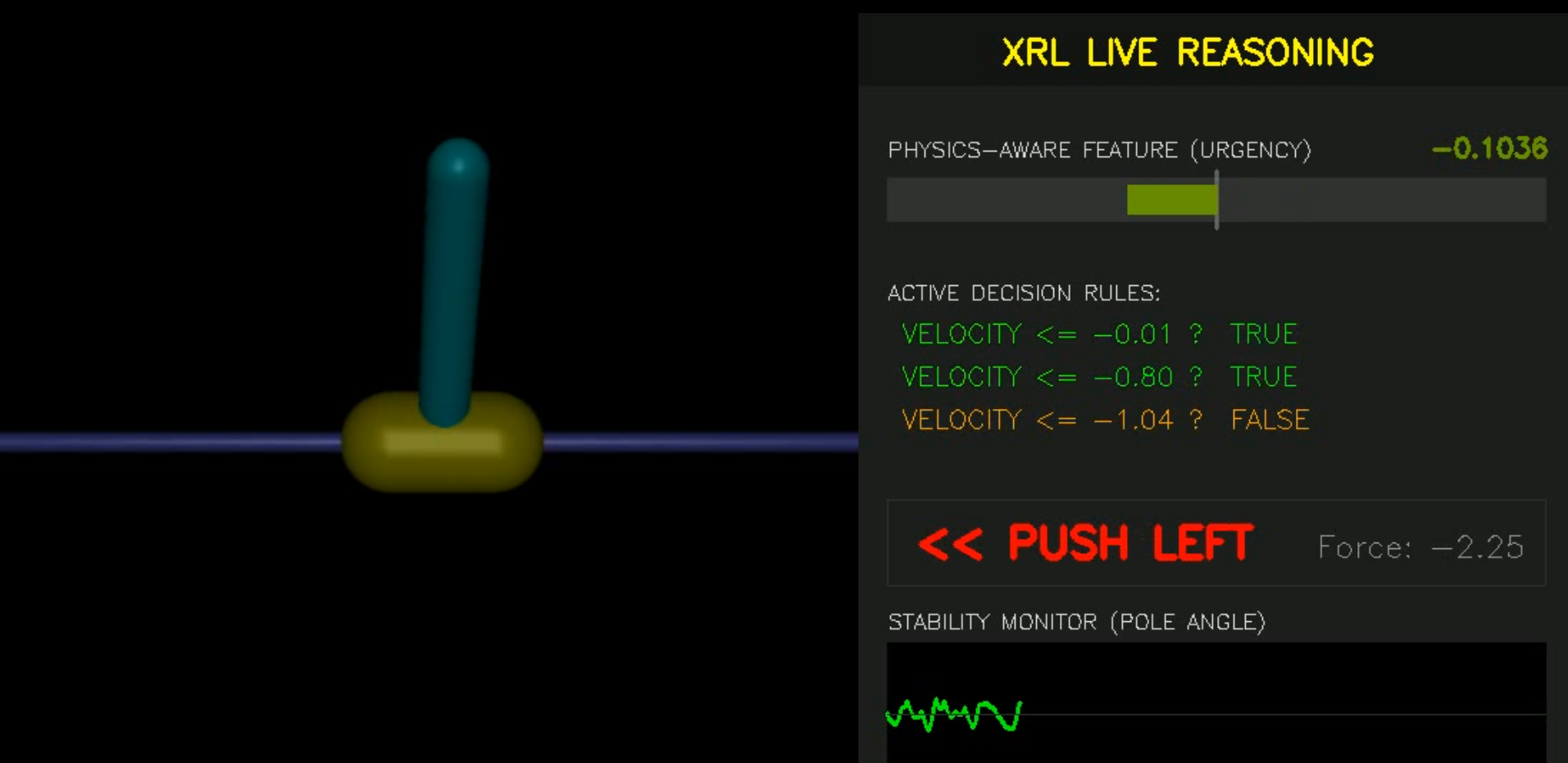}
    \caption{Live Decision Path Visualization. This real-time diagnostic dashboard provides local 
interpretability by simultaneously displaying the physical simulation (left) alongside the active 
IF-THEN decision path, the engineered urgency metric, and a live stability limit-cycle monitor 
(right). }
    \label{fig:placeholder}
\end{figure}

\section{discussion}
\subsection{Strength of the Approach}
The primary strength of this methodology is the achievement of 
a Distillation of a complex neural network teacher into a simpler, interpretable equivalent student. By utilizing "Noisy Oracle Rollouts," the 
dataset captured critical recovery behaviors that a standard optimal 
agent would never exhibit. Furthermore, the introduction of the 
physics-aware Pole\_Urgency feature allowed the Decision Tree to 
approximate non-linear control logic (diagonal decision boundaries) 
with a significantly shallower depth (max depth = 7), maintaining 
high performance without sacrificing human readability. 

\subsection{Limitations}
While the Inverted Pendulum is a foundational benchmark for control 
theory, the primary limitation of this approach is the inherent Bang
Bang control strategy of Decision Trees. As visualized in the Control 
Effort analysis, the tree's piecewise constant output causes high
frequency mechanical chatter. While capable of consistently balancing the system , this 
jittery control signal could induce severe mechanical wear and tear if 
applied directly to physical hardware, contrasting sharply with the 
smooth outputs of a neural network. 

\subsection{Generalization to Real-World Systems}
The principles demonstrated here highlight potential pathways for scaling interpretable control to complex autonomous systems. While scaling to high-dimensional environments presents significant challenges, in Autonomous Driving, for example, a neural 
network could handle complex perception, while a distilled Decision 
Tree could serve as an interpretable "Safety Layer" for path planning. 
This would allow human auditors to verify exactly why a vehicle chose 
to brake or swerve. Similarly, in Robotics, this approach facilitates the 
transition from "Black-Box" AI to "Certifiable AI," where control 
logic must be legally and technically defensible.

\vspace{1cm}

\section{conclusion and future work}
\vspace{1em}

\textit{Conclusion:}
\vspace{0.5em}

This work successfully demonstrated the viability of Explainable Reinforcement Learning (XRL) via Policy Distillation. We transitioned an opaque TD3 agent into a transparent, depth-limited Decision Tree without sacrificing the 100\% task success rate. By integrating control theory metrics—specifically Bimodal Limit Cycles, Control Quantization, and Action Chattering—we provided a rigorous framework for validating surrogate models. This demonstrates that a distilled policy can maintain strict Bounded-Input Bounded-Output (BIBO) stability while operating entirely on human-readable IF-THEN rules.

\vspace{1em}
\textit{Future Work:}
\vspace{0.5em}

Future research should focus on mitigating the mechanical limitations of discrete surrogate models. Exploring output smoothing techniques (such as passing the tree's output through a low-pass filter) could reduce the observed action chattering. Additionally, extending this distillation pipeline to higher-dimensional continuous control environments (e.g., multi-joint robotic locomotion) and validating the "Interpretable Safety Layer" concept on physical hardware remains a critical next step toward safe autonomous operation.

\end{document}